%% file: model_comparison.tex
\documentclass[]{article}
\usepackage{proceed2e}
\usepackage{graphicx}
\usepackage{amsmath, amsfonts}
\usepackage{tikz}

\usepackage[authoryear]{natbib}
\usepackage{hyperref}
\usepackage{amsthm}
\usepackage{tkz-graph}
\usepackage{url}

\usetikzlibrary{arrows}
\pgfarrowsdeclare{biggertip}{biggertip}{%
  \setlength{\arrowsize}{1pt}  
  \addtolength{\arrowsize}{.5\pgflinewidth}  
  \pgfarrowsrightextend{0}  
  \pgfarrowsleftextend{-5\arrowsize}  
}{%
  \setlength{\arrowsize}{1pt}  
  \addtolength{\arrowsize}{.5\pgflinewidth}  
  \pgfpathmoveto{\pgfpoint{-5\arrowsize}{4\arrowsize}}  
  \pgfpathlineto{\pgfpointorigin}  
  \pgfpathlineto{\pgfpoint{-5\arrowsize}{-4\arrowsize}}  
  \pgfusepathqstroke  
}  
\tikzset{
  LabelStyle/.style = { rectangle, rounded corners, draw,
                        minimum width = 2em,
                        text = red, font = \bfseries },
  VertexStyle/.append style = { inner sep=5pt,
                                font = \Large\bfseries},
  EdgeStyle/.append style = {->, arrowhead=1em} }

\theoremstyle{definition}

\usepackage{times}

\title{Probabilistic Models for Computerized Adaptive Testing: Experiments}

\author{ {\bf Martin~Plajner }\\
Institute of Information Theory and Automation\\
Academy of Sciences of the Czech Republic\\
Pod vod\'{a}renskou v\v{e}\v{z}\'{i} 4\\
Prague 8, CZ-182 08\\
Czech Republic\\
\And
{\bf Ji\v{r}\'{i}~Vomlel}  \\
Institute of Information Theory and Automation\\
Academy of Sciences of the Czech Republic\\
Pod vod\'{a}renskou v\v{e}\v{z}\'{i} 4\\
Prague 8, CZ-182 08\\
Czech Republic\\
}

\begin{document}

\maketitle

\begin{abstract}
This paper follows previous research we have already performed in the area of Bayesian networks models for CAT. We present models using Item Response Theory (IRT -  standard CAT method), Bayesian networks, and neural networks. We conducted simulated CAT tests on empirical data. Results of these tests are presented for each model separately and compared.\end{abstract}

\input{"mc_intro"}
\input{"mc_sec0"}
\input{"mc_sec1"}
\input{"mc_sec2"}
\input{"mc_sec3"}

\input{"mc_sec4"}

\subsubsection*{Acknowledgements}
The work on this paper has been supported from GACR project n. 16-12010S.

\renewcommand{\refname}{\normalfont\selectfont\normalsize\bfseries References} 
\bibliographystyle{apalike}
\bibliography{model_comparison}

\end{document}

%% file: mc_intro.tex
\section{INTRODUCTION}
All of us are in touch with different ability and skill checks almost every day. The computerized form of testing is also getting an increasing attention with the spread of computers, smart phones and other devices which allow easy contact with target groups. This paper focuses on the Computerized Adaptive Testing (CAT)~\citep{Linden2000,Almond1999,Almond2015} and it follows a previous research paper~\citep{Plajner2015}. 

In this previous paper we explained the concept of CAT. Next, we describe our empirical data set. The use of Bayesian networks for CAT was discussed and we constructed different types of Bayesian network models for CAT. These models were tested on empirical data. The results were presented and discussed.

In this paper we present two additional model types for CAT: Item Response Theory (IRT) and neural networks. Moreover, new BN models are proposed in this paper. We conducted simulated CAT tests on the same empirical data as in the previous paper. This allows us to make comparisons of two new model types (BN and NN) with the CAT standard IRT model. Results are presented for each model separately and then they are all compared.

%% file: mc_sec0.tex
\section{CAT PROCEDURE AND MODEL EVALUATION}
\label{sec:intro}
All models proposed in this paper are supposed to serve for adaptive testing. In this section we briefly outline the process of adaptive testing\footnote{Additional information about CAT can be found in~\citep{Wainer2015}} with the help of these models and methods for their evaluation. For every model we used similar procedures. The specific details for each model type are discussed in the corresponding sections. At this point we discuss the common aspects.

In every model type we have the following types of variables. For some models they have a different specific name because of an established naming convention of the corresponding method. Nevertheless, the meaning of these variables is the same and we explain differences for each model types. In this paper we use two types of variables:
\begin{itemize} 
\item A set of $n$ variables we want to estimate $\mathcal{S} = \{S_1,\ldots,S_n\}$. These variables represent latent skills (abilities, knowledge) of a student.
We will call them skills or skill variables. We will use symbol $\boldsymbol{S}$ to denote the multivariable $\boldsymbol{S} = (S_1,\ldots,S_n)$ taking states $\boldsymbol{s} = (s_{1,i_1},\ldots,s_{n,i_n})$. 
\item A set of $p$ questions $\mathcal{X} = \{X_1,\ldots,X_p\}$.  
We will use the symbol $\boldsymbol{X}$ to denote the multivariable $\boldsymbol{X} = (X_1,\ldots,X_p)$ taking states $\boldsymbol{x} = (x_1,\ldots,x_p)$.
\end{itemize}

We collected data from paper tests conducted by grammar schools' students. The description of the test and its statistics can be found in the paper~\citep{Plajner2015}. All together, we have obtained 281 test results. Experiments were performed with each model of each type that is described in following sections. We used 10-fold cross-validation method. We learned each model from $\frac{9}{10}$ of randomly divided data. 
The remaining $\frac{1}{10}$ of the data set served as a testing set. 
This procedure was repeated 10 times to obtain 10 learned student models with the same structure and different parameters. 

With these learned models we simulate CAT using test sets. For every student in a test set a CAT procedure consists of the following steps:
\begin{itemize}
	\item The next question to be asked is selected.
	\item This question is asked and an answer is obtained.
	\item This answer is inserted into the model.
	\item The model (which provides estimates of the student's skills) is updated.
	\item (optional) Answers to all questions are estimated given the current estimates of student's skills.
\end{itemize}
This procedure is repeated as long as necessary. It means until we reach a termination criterion, which can be, for example, a time restriction, the number of questions, or a confidence interval of the estimated variables. Each of these criteria would lead to a different learning strategy~\citep{Vomlel2004}, but finding a global optimal selection with these strategies would be NP-hard~\citep{Lin2005}. We have chosen an heuristic approach based on greedy optimization methods. Methods of the question selection differ for each model type and are explained in the respective sections. All of them use the greedy strategy to select questions. 

To evaluate models we performed a simulation of CAT test for every model and for every student. During testing we first estimated the skill(s) of a student based on his/her answers. Then, based on these estimated skills we used the model to estimate answers to all questions $\boldsymbol{X}$. Let the test be in the step $s$ ($s-1$ questions asked). At the end of the step $s$ (after updating the model with a new answer) we compute marginal probability distributions for all skills $\boldsymbol{S}$. Then we use this to compute estimations of answers to all questions, where we select the most probable state of each question $X_i \in \mathcal{X}$:
$$
x_i^* = \arg\max_{x_i}{P(X_{i}=x_i|\boldsymbol{S})}.
$$
By comparing this value to the real answer to $i-th$ question $x_i'$ we obtain a success ratio of the response estimates for all questions $X_i \in \mathcal{X}$ of a test result $t$ (particular student's result) in the step $s$
\begin{eqnarray*}
\operatorname{SR}_s^t & = & \frac{\sum_{X_i \in \mathcal{X}}{I(x_{i}^* = x_{i}')}}{|\mathcal{X}|} \ , \ \mbox{where}\\
I(expr) & = & \left\{ \begin{array}{ll}
1 & \mbox{if $expr$ is true}\\
0 & \mbox{otherwise.}
\end{array}\right.
\end{eqnarray*}
The total success ratio of one model in the step $s$ for all test data is defined as
\begin{eqnarray*}
\operatorname{SR}_s & = & \frac{\sum_{t=1}^N{\mathrm{SR}_s^t}}{N} \enspace .
\end{eqnarray*}
$\operatorname{SR}_0$ is the success rate of the prediction before asking any questions.

%% file: mc_sec1.tex
\section{ITEM RESPONSE THEORY}
The beginning of Item Response Theory (IRT) stem back to 5 decades ago and there is a large amount of resources available, for example,~\citep{Lord1968, Rasch1960, Rasch1993}. IRT allows more specific measurements of certain abilities of an examinee. It expects a student to have an ability (skill) which directly influences his/her chance of answering a question correctly. When we have only one variable\footnote{There are variants of multidimensional IRT model where it is possible to have more then one ability but in this section we are going to discuss only models with one only.}, it is common to refer to it as a proficiency variable. This ability is called latent ability or a latent trait $\theta$. The trait $\theta$ corresponds to the general skill $S_1$ defined in the Section~\ref{sec:intro}. Every question of the IRT model has an associated item response function (IRF) which is a probability of a successful answer given $\theta$. 

We fitted our data on the 2 parametric IRT model. It means that characteristic Item Response Functions, as the probability of a correct answer to \textit{i-th} given the ability $\theta$, are computed by the formula
$$
p_i(\theta) = \frac{1}{1+e^{-a_i(\theta-b_i)}}
$$
where $a_i$ sets the scale of the question (this sets its discrimination ability - a steeper curve better differentiate between students), $b_i$ is the difficulty of the question (horizontal position of a curve in space). 

For question selection step of CAT we use item information of a question $i$ that it is given by the formula
$$I_i(\theta)=\frac{(p_i'(\theta))^2}{p_i(\theta)q_i(\theta)}$$
where $p_i'$ is the derivation of the item response function $p_i$. This item information provides one, and most straightforward, way of the next question selection. In every step the question $X^*$ which is selected is one with the highest item information. 
$$X^*(\theta) = \arg\max_i I_i(\theta)$$
This approach minimizes the standard error of the test procedure~\citep{Hambleton1991} because the standard error of measurement $SE_i$ produced by $i-th$ item is defined as
$$SE_i(\theta) = \frac{1}{\sqrt{I_i(\theta)}}.$$
This means that the better precision of difficulty we are able to achieve while asking questions the smaller error of measurement.

\begin{figure*}[htbp]
\centering
\includegraphics[width=0.9\textwidth, height=0.45\textwidth]{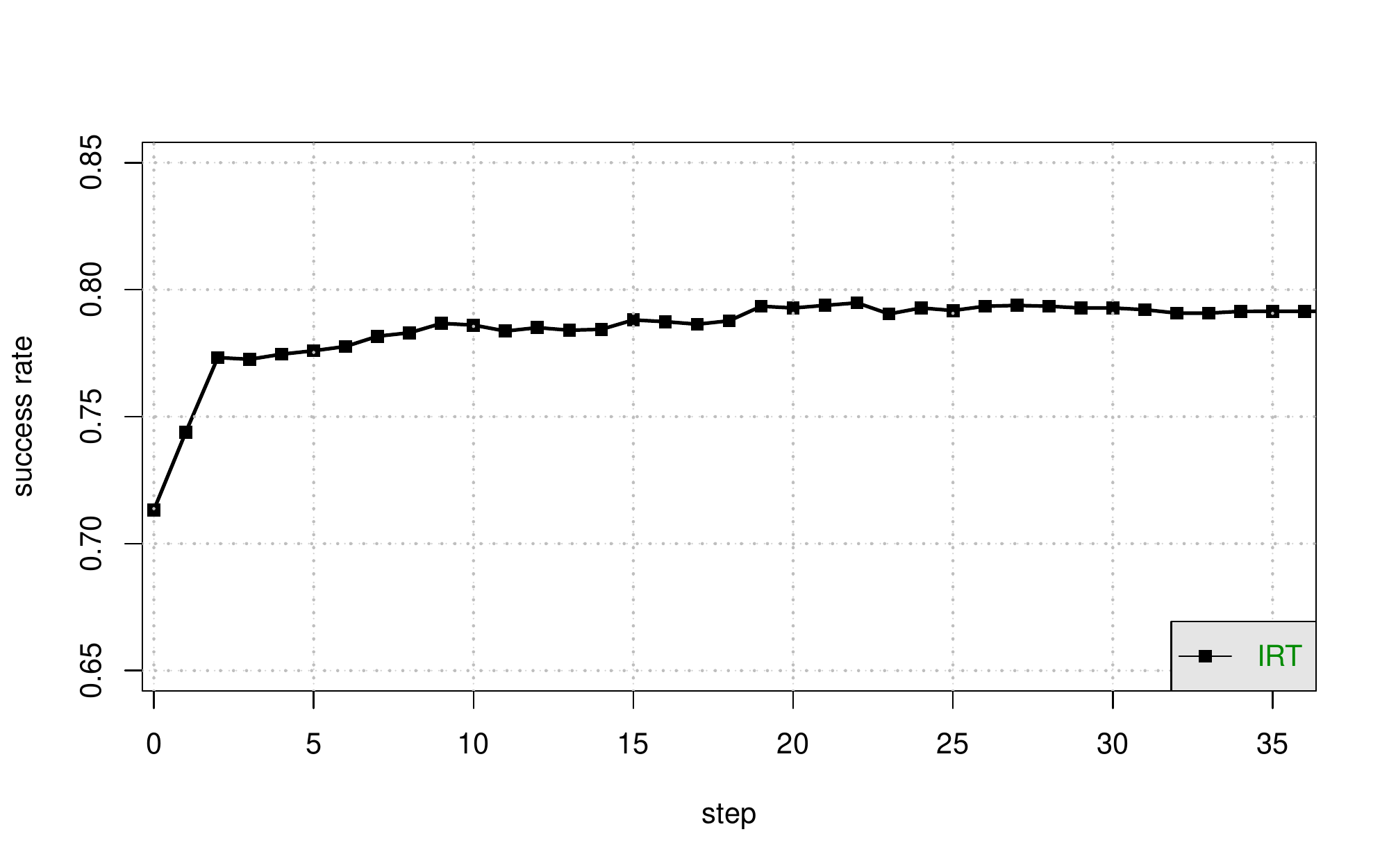}
\caption{Success rates of IRT model}
\label{fig:res_IRT}
\end{figure*}

The result of CAT simulation is displayed in the Figure~\ref{fig:res_IRT}. We can notice that this model is able to choose correct questions to ask very quickly and its prediction success rises after asking the first two. After these questions it can not improve much any more. This is caused by the simplicity of the model.

%% file: mc_sec2.tex
\section{BAYESIAN NETWORKS}
In this section we use Bayesian networks (BN) as CAT models. Details about BNs can be found in~\citep{Jensen2007, Kjrulff2008}. The use of BNs in educational assessment is discussed in~\citep{Almond2015, Culbertson2015, Millan2010}. This topic is also discussed, for example, in~\citep{Vomlel2004a, Vomlel2004}.

A Bayesian network is a probabilistic graphical model, a structure representing conditional independence statements. It consists of the following: 
\begin{itemize}
	\item a set of variables (nodes),
	\item a set of edges,
	\item a set of conditional probabilities.
\end{itemize}
Specific details about the use of BNs for CAT can be found in~\citep{Plajner2015}. Types of nodes in our BNs correspond to types of variables defined in the Section~\ref{sec:intro}. In this paper we use question nodes with only Boolean states, i.e., question is either correct or incorrect. Edges are defined usually between skills and questions (we present examples of connections in figures). Conditional probability values have been learned using standard EM algorithm for BN learning.

In this paper we use a modified method for model scoring compared to the method used in our previous research. The current method is described in the section~\ref{sec:intro}. The difference is that in this case we estimate answers to all questions in the question pool and then compare to real answers in every step. In the previous version we were estimating answers only to unanswered questions in every step. It led to a skewed results interpretation because the value in the denominator of the success rate
$$\operatorname{SR}_s^t  =  \frac{\sum_{X_i \in \mathcal{X}}{I(x_{i}^* = x_{i}')}}{|\mathcal{X}|} $$
was decreasing in every step. The modified version is comparing all questions and because of that the denominator stays the same in every step.

\begin{figure*}[htb]
\centering
\includegraphics[width=0.7\textwidth]{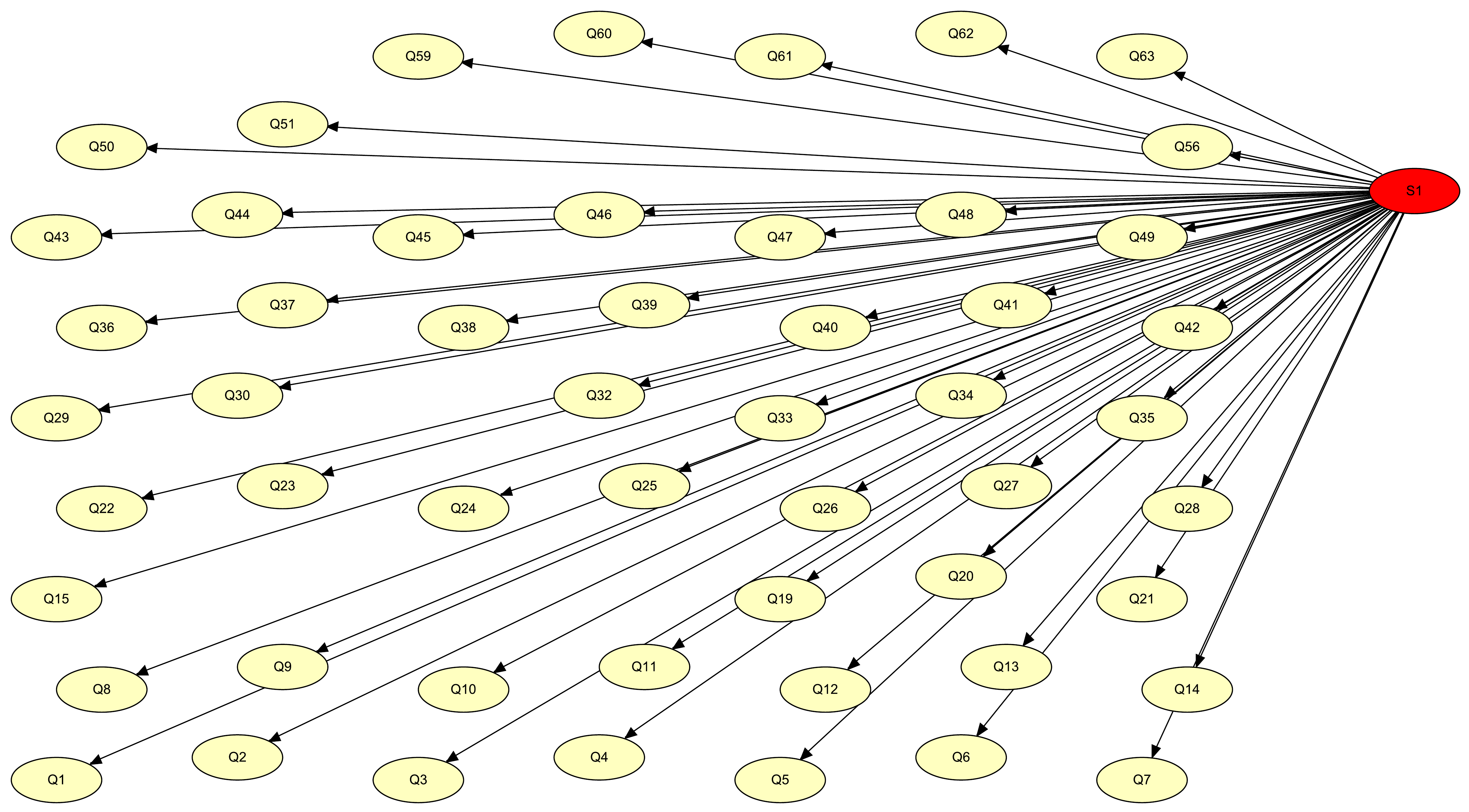}
\caption{Bayesian network with one hidden variable and personal information about students}
\label{fig:net_one_skill}
\end{figure*}

\begin{figure*}[htb]
\centering
\includegraphics[width=0.7\textwidth]{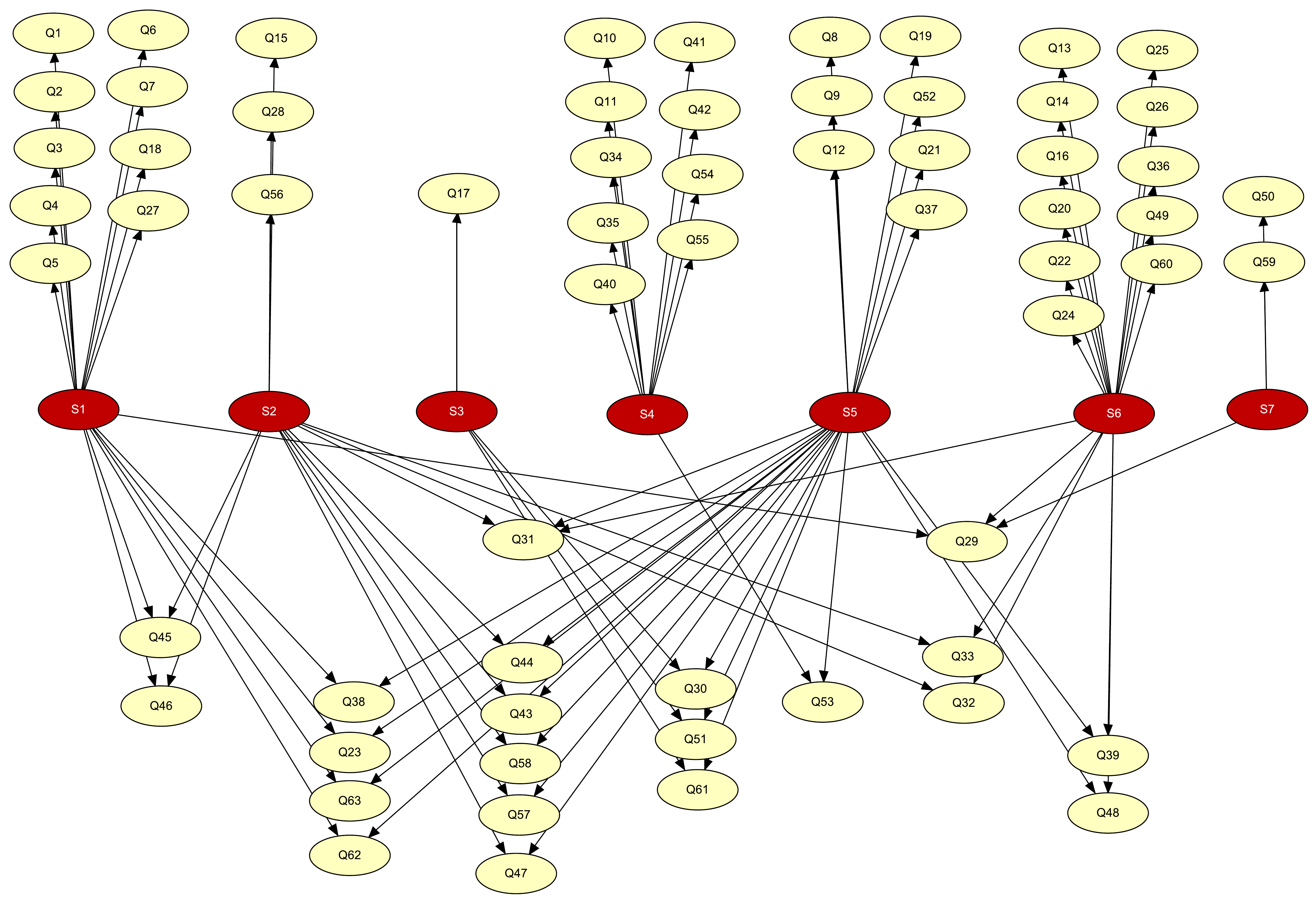}
\caption{Bayesian network with 7 hidden variables (the old expert model)}
\label{fig:expert_old}
\end{figure*}

From previous models we selected the model marked as ``b3'' and ``expert''. The former means that it has Boolean answer values, there is one skill variable having 3 states and no additional information (personal data of students) was used. See Figure~\ref{fig:net_one_skill} for its structure. The later is an expert model with 7 skill nodes (each having 2 states), Boolean answer values and no additional information about students was used. See Figure~\ref{fig:expert_old} for its structure. 

\begin{figure*}[htb]
\centering
\includegraphics[width=0.7\textwidth]{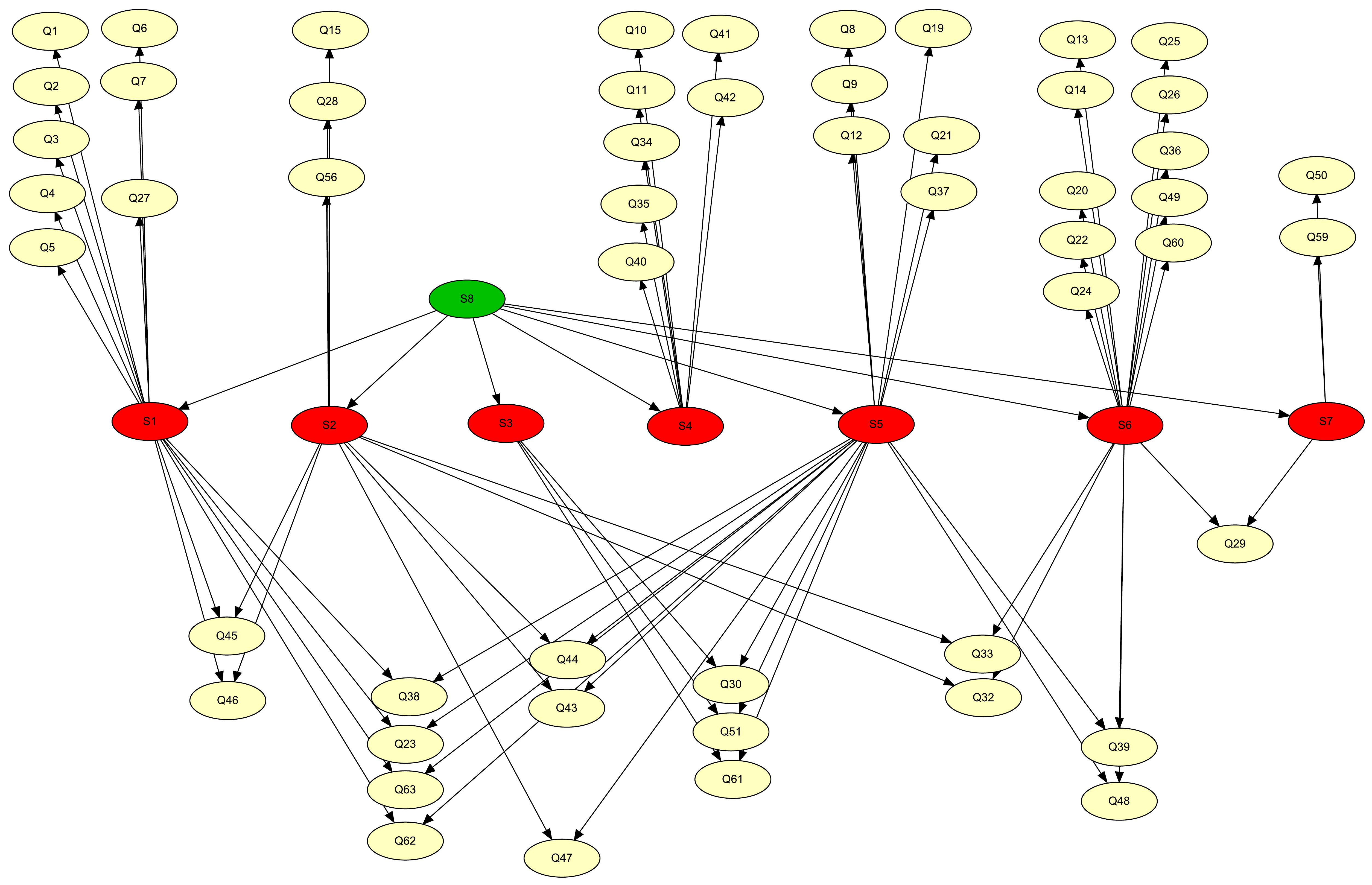}
\caption{Bayesian network with 7+1 hidden variables (the new expert model)}
\label{fig:expert_new}
\end{figure*}

In this paper we present three new BN models. The first two are modifications of ``b3'' model. They have the same structure and differ only in the number of states of their skill node. We present experiments with 4 and 9 states. We performed experiments with different numbers of states as well, but they do not provide more interesting results. Next, we add a modified expert model. This modified model has also Boolean questions and no additional information. We have added one state to 7 skill nodes from the previous version (they have 3 states in total now). The reason for this addition is an analysis of the question selection criterion. We select questions by minimizing the expected entropy over skill nodes. With only two states it means that we are pushing a student into one or the other side of the spectrum (basically, we want him to be either good or bad). With 3 states we allow them to approach mediocre skill quality as well. Moreover, we realized that the model structure as in the Figure~\ref{fig:expert_old} has only skills that are very specialized. We introduce a new 8th skill node which connect previous 7 skill nodes. Its representation is an overall mathematical skill combining all other skills. It allows skills on the lower level to influence each other and to provide evidence between themselves. The final model structure is in the Figure~\ref{fig:expert_new}.

\begin{figure*}[htb]
\centering
\includegraphics[width=0.8\textwidth]{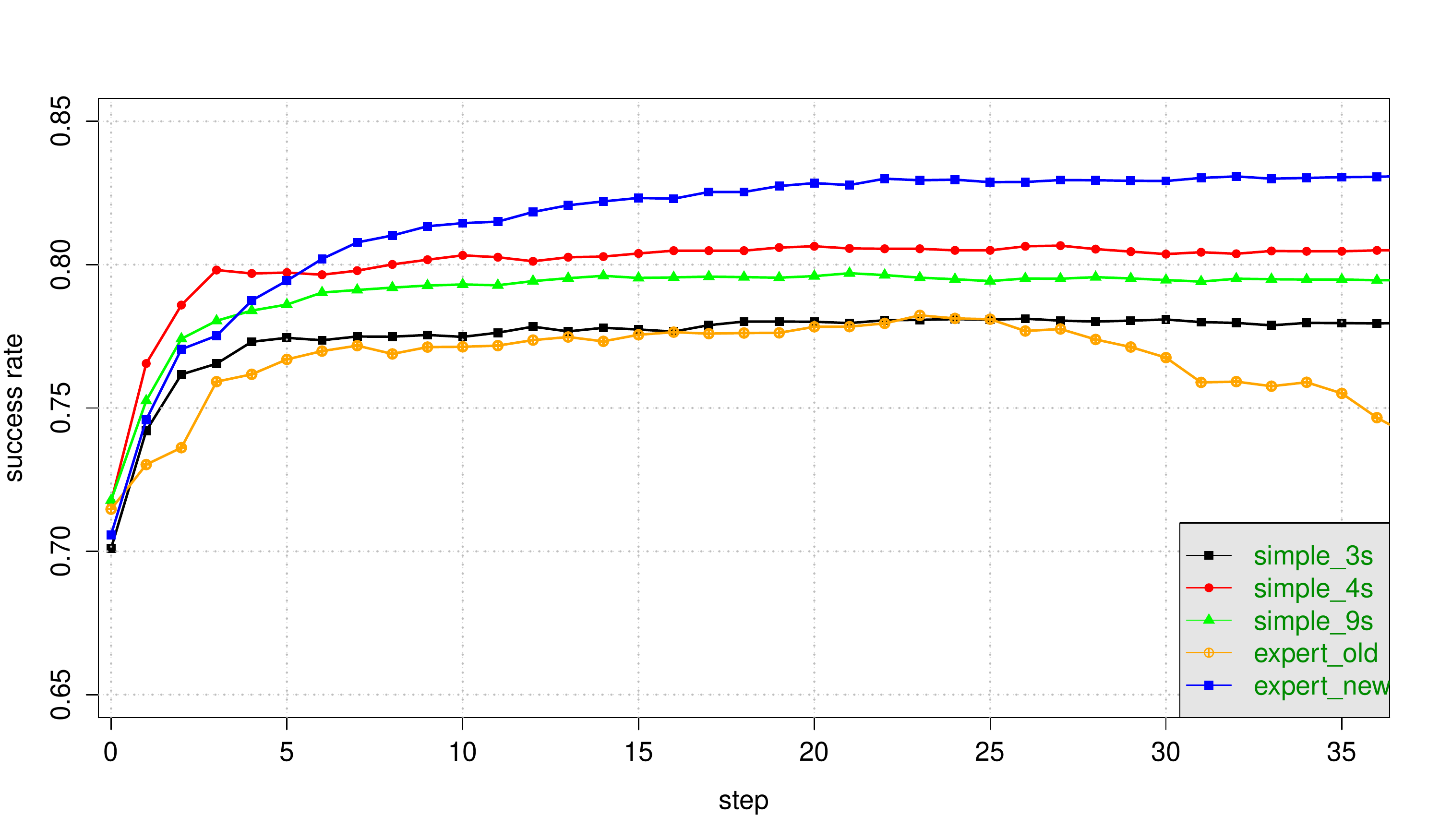}
\caption{Results of CAT simulation with BNs}
\label{fig:res_expert}
\end{figure*}

All models are summarized in the Table~\ref{tab:BayesianNetworkModelsOverview}. Results of CAT simulation with BN models are displayed in the Figure~\ref{fig:res_expert}. Increasing the number of states of one skill node improved prediction accuracy of the model (simple\_4s, simple\_9s), but only slightly. As we can see, one additional state (4 states in total) is better than more states (9). This confirms our expectation that simply adding node states can not improve the model quality for long due to over fitting of the model. Next, we can observe that there is a large difference between the new and the old expert model. The success rate of the new version exceeds all other models. Adding additional skill node connecting other skills proved to be a correct step. Possibilities in the model structure are still large and it remains to be explored how to create the best possible structure.

\begin{table}
	\centering
		\begin{tabular}{lccc}
			Model name & Figure & \rotatebox{90}{No. of skill nodes} & \rotatebox{90}{No. of states of skill nodes} \\
			\hline
			simple\_3s & \ref{fig:net_one_skill} &  1 & 3 \\
			simple\_4s & \ref{fig:net_one_skill} &  1 & 4 \\ 			
			simple\_9s & \ref{fig:net_one_skill} &  1 & 9 \\ 
			expert\_old & \ref{fig:expert_old} & 7 & 2  \\
			expert\_new & \ref{fig:expert_new} & 7+1 & 3 \\ 			
			\hline
			\end{tabular}
	\caption{Overview of Bayesian network models}
	\label{tab:BayesianNetworkModelsOverview}
\end{table}

%% file: mc_sec3.tex
\section{NEURAL NETWORKS}
Neural networks are models for approximations of non-linear functions. For more details about NNs, please refer to~\citep{haykin2009, aleksander1995}. 

There are three different parts of a NN:
\begin{enumerate}
	\item an input layer,
	\item several hidden layers, and
	\item an output layer.
\end{enumerate}

We use NN as a student model. We feed student answers to the input layer. These values are transformed to the hidden layer(s). There is no general rule how to choose the number of hidden layers and their size. In our case we performed experiments with one hidden layer of different sizes. The hidden layer then further transforms to the output layer. NNs are not suitable for unsupervised learning. Because of that, we do not estimate an unknown student skill in the output layer. We would not have any target value needed during the learning step of the NN. Instead of that, we estimate the score (the test result) of a student directly. The score of a student is known for every student at the time of learning. The output layer then provide an estimate of this score. Nevertheless, this score is a corresponding variable to skill variables described in the Section~\ref{sec:intro}

To select the next question we use the following procedure. We want the selected question to provide us as much information as possible about the tested student. That means that a student who answers incorrectly should be as far as possible on the score scale from another who answers correctly. Let the $S|X_{i,x}$ be the score prediction after answering the $i-th$ question's state $x$, $P(X_{i,x})$ the probability of state $x$ to be the answer to the question $i$. $P(X_{i,x})$ can be obtained, for example, by statistical analysis of answers. We select a question $X^*$ maximizing the variance of predicted scores:

\begin{align*}
X^* &= \arg\max_i \mathrm{V\underset{x}ar} (SC|X_{i,x}) \\ &= \sum_x P(X_{i,x})(SC|X_{i,x} - \overline{SC|X_i}) , \text{where} \\
\overline{SC|X_i} &= \sum_xP(X_{i,x}){SC|X_{i,x}}
\end{align*}
is the mean value of predicted scores.

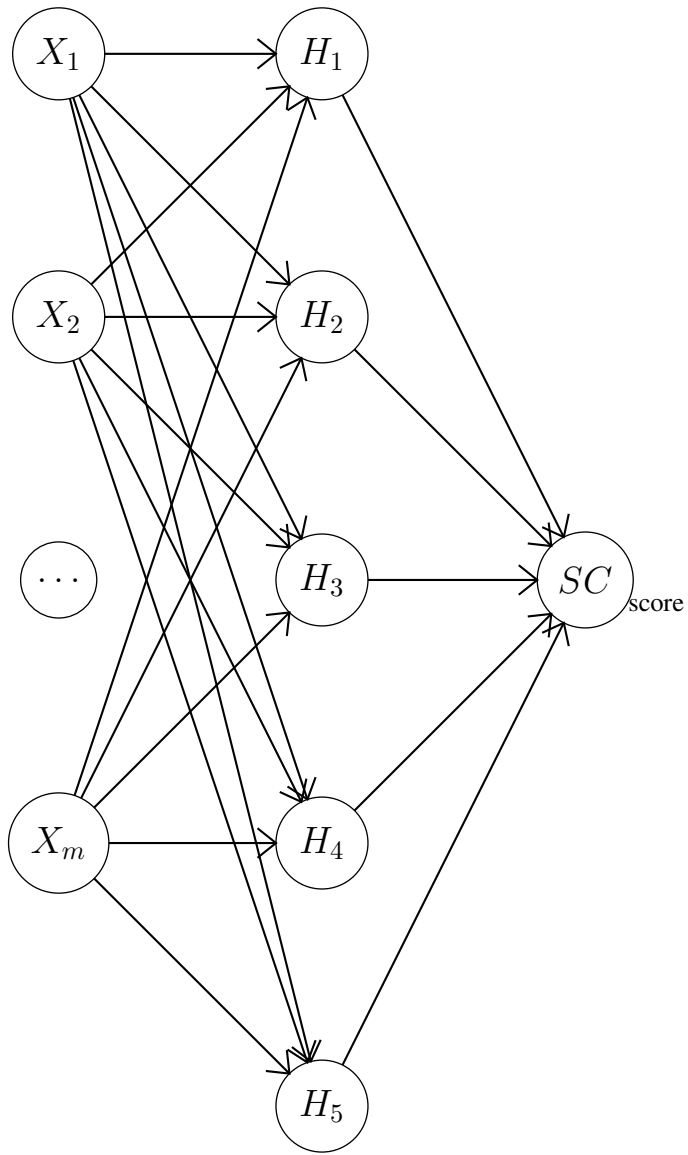
\begin{figure*}%
\centering
\begin{tikzpicture}
  \SetGraphUnit{3.5}
	\SetUpEdge[style={->,>=biggertip}]
	\Vertex[Math,L=X_1] {X1}
	\SO[Math,L=X_2](X1){X2} 
	\SO[Math,L=\ldots](X2){X3} 
	\SO[Math,L=X_m](X3){XM} 
	
	\EA[Math,L=H_3](X3){H3}
\NO[Math,L=H_2](H3){H2}
\NO[Math,L=H_1](H2){H1}
\SO[Math,L=H_4](H3){H4}
\SO[Math,L=H_5](H4){H5}

  \EA[Math,L=SC](H3){SC}
	\node	at ([shift=(-7cm:-1cm)]SC) (l) {score};

\Edge(X1)(H1)
\Edge(X1)(H2)
\Edge(X1)(H3)
\Edge(X1)(H4)
\Edge(X1)(H5)

\Edge(X2)(H1)
\Edge(X2)(H2)
\Edge(X2)(H3)
\Edge(X2)(H4)
\Edge(X2)(H5)

\Edge(XM)(H1)
\Edge(XM)(H2)
\Edge(XM)(H3)
\Edge(XM)(H4)
\Edge(XM)(H5)

\Edge(H1)(SC)
\Edge(H2)(SC)
\Edge(H3)(SC)
\Edge(H4)(SC)
\Edge(H5)(SC)
	
\end{tikzpicture}
\caption{Neural network with 5 hidden neurons}%
\label{fig:NN}
\end{figure*}

\begin{figure*}[htb]
\centering
\includegraphics[width=0.8\textwidth]{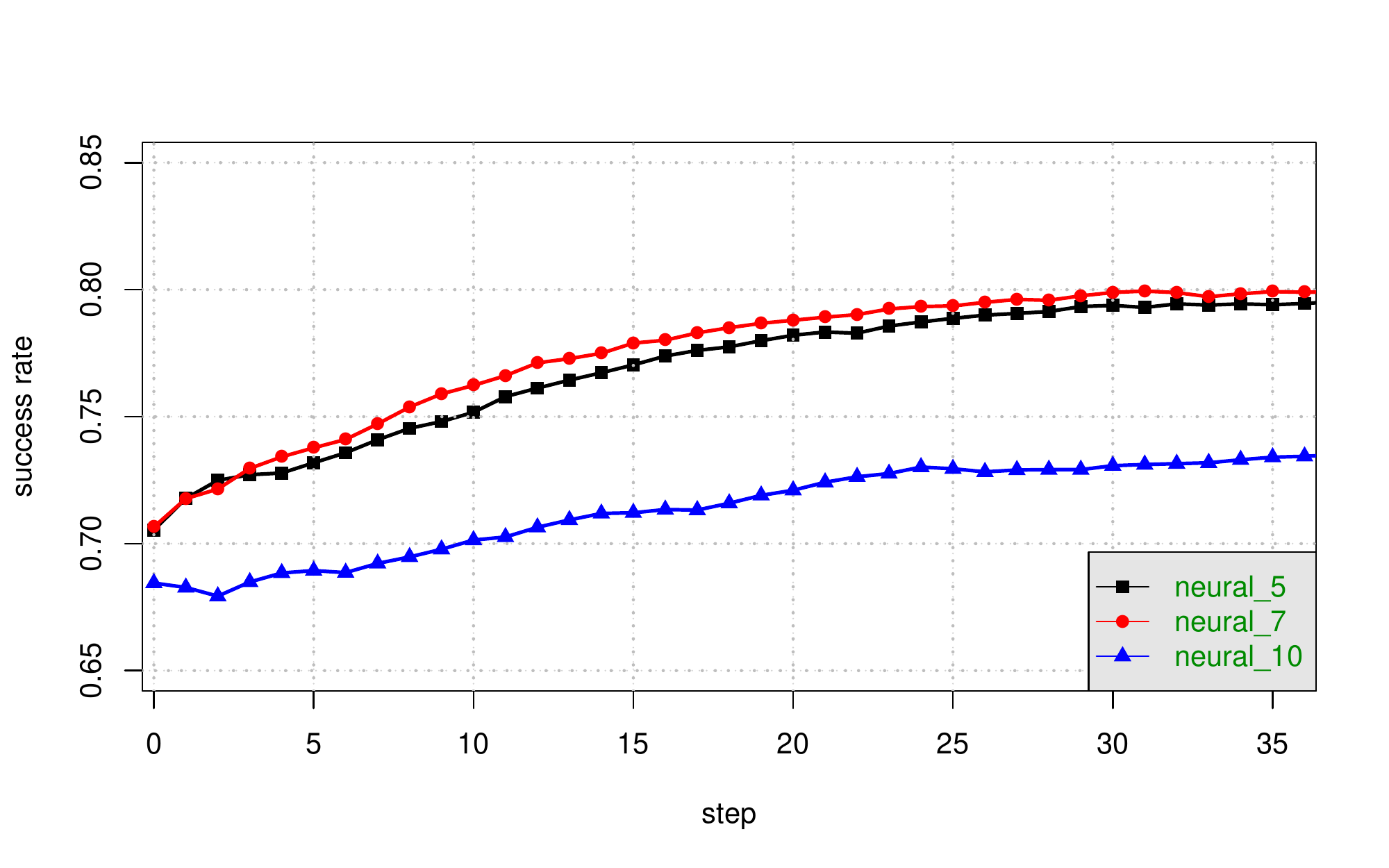}
\caption{Results of CAT simulation with NNs}
\label{fig:res_neural}
\end{figure*}

In our experiment we used only one hidden layer with many different numbers of hidden neurons. From them we select models with 3, 5, and 7 neurons in the hidden layer because they provide the most interesting results. The structure of the network with 5 hidden neurons is in the Figure~\ref{fig:NN}. Results of CAT simulation with NN models are displayed in the Figure~\ref{fig:res_neural}. As we can see in this figure, the quality of estimates while using NNs increases very slowly. This may be caused by the question selection criterion. If we were selecting better questions, it is possible that the success rate would be increasing faster. It remains to be explored which selection criterion would provide such questions. Nevertheless, this better question selection does not change the final prediction power of the model (the maximal success rate would not be exceeded). This prediction power could be increased by using a different version of NNs. More specifically, we will perform experiments with one of the recurrent versions of NNs, i.e., Elman's networks or Jordan's networks.

%% file: mc_sec4.tex
\section{MODEL COMPARISON AND CONCLUSIONS}
\begin{figure*}[htb]
\centering
\includegraphics[width=0.8\textwidth]{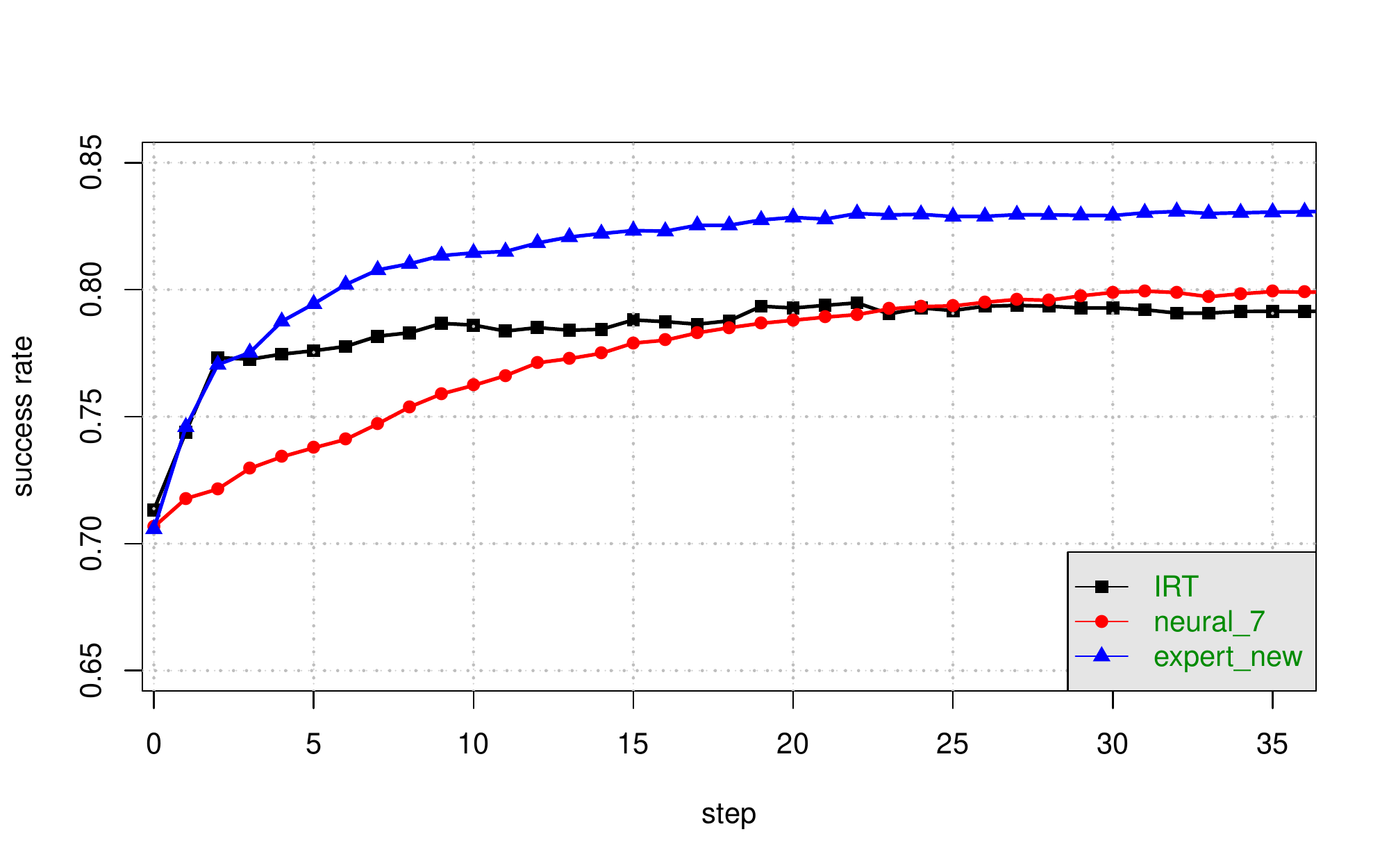}
\caption{CAT simulation results comparison}
\label{fig:all_res}
\end{figure*}

We present a graphical comparison of all three model types in the Figure~\ref{fig:all_res}. One model is selected from each type. We can see that the neural network model scored the worst result. This may be further improved by a better NN structure and better question selection process. The new BN expert model is scoring the best. Even in this case we believe that further improvements are possible to increase its success rate. We will focus our future research into methods for BN models creation and criteria for their comparison. Especially, we would like to use a concept of the local structure in BN models~\citep{Diez2007}. That would allow us to create more complex models, yet with less parameters to be estimated during learning. Both previous models can be compared with the IRT model which is the standard in the field of CAT.

